\pdfoutput=1
\documentclass{article}
\PassOptionsToPackage{numbers, compress}{natbib}
\usepackage[final]{neurips_2023}

\usepackage[utf8]{inputenc} % allow utf-8 input
\usepackage[T1]{fontenc}    % use 8-bit T1 fonts
\usepackage{hyperref}       % hyperlinks
\usepackage{url}            % simple URL typesetting
\usepackage{booktabs}       % professional-quality tables
\usepackage{nicefrac}       % compact symbols for 1/2, etc.
\usepackage{microtype}      % microtypography
\usepackage{xcolor}         % colors

\usepackage{latexsym}
\usepackage{inconsolata}
\usepackage{multirow}
\usepackage{bbding}

\usepackage{comment}
\usepackage{amsmath,amssymb,amsfonts}

\usepackage[ruled]{algorithm}
\usepackage{algpseudocode}
\usepackage{caption}
\usepackage{xfrac}

\usepackage{comment}
\usepackage[capitalize]{cleveref}
\crefname{section}{Sec.}{Secs.}
\Crefname{section}{Section}{Sections}
\Crefname{table}{Table}{Tables}
\crefname{table}{Tab.}{Tabs.}
\let\oldnl\nl% Store \nl in \oldnl
\newcommand{\nonl}{\renewcommand{\nl}{\let\nl\oldnl}}% Remove line number for one line
\def \xyes {\CheckmarkBold}
\def \xno {\XSolidBrush} 

\def \Padding {RandomSampling+Padding}
\def \GroupByLength {GroupByLength+Padding}
\def \PaddingFree {MiniBatchPacking+PosIDd}
\def \Packing {FixedLengthPacking}
\def \PackingPosId {FixedLengthPacking+PosID}
\def \Multipack {Multipack+PosID}
\def \Sortedpack {SortedPacking+PosID}
\def \Randompack {RandomPacking+PosID}

\def \X {\mathbf{X}}
\def \x {\mathbf{x}}
\title{Enhancing Training Efficiency Using Packing with Flash Attention}

\author{
Achintya Kundu \\
IBM Research \\
\texttt{ achintya.k@ibm.com } 
\\ \And
Rhui Dih Lee \\
IBM Research \\
\texttt{ rhui.dih.lee@ibm.com }
\\ \AND
Laura Wynter \\
IBM Research \\
\texttt{ lwynter@sg.ibm.com }
\\ \And
Raghu Kiran Ganti \\
IBM Research \\
\texttt{ rganti@us.ibm.com }
\\ \And
Mayank Mishra \\
IBM Research \\
\texttt{ mayank.mishra2@ibm.com}
}

\begin{document}
\maketitle

%%%%%%%%%%%%%====== Abstract  ==============
\begin{abstract}
Padding is often used in tuning LLM models by adding special tokens to shorter training examples to match the length of the longest sequence in each batch. While this ensures uniformity for batch processing, it introduces inefficiencies by including irrelevant padding tokens in the computation and wastes GPU resources. Hugging Face  trainer has always offered the option to use packing to combine multiple training examples, allowing for maximal utilization of GPU resources. However, up till now, it did not offer  proper masking of each packed training example. This capability has now been added to Hugging Face Transformers 4.44. We   analyse this new feature and show the benefits across different variations of packing.
\end{abstract}

%%%%%----------------------------------------------------------------------------------------------------------
%%%%%----------------------------------------------------------------------------------------------------------
%%%%-----------------------------------
\section{Introduction}
%%%%-----------------------------------

In LLM tasks, data is represented as sequences of tokens, where each token typically corresponds to a word, character, or sub-word. These sequences form the input to the models, which aim to learn meaningful representations and capture intricate patterns within the data. However, the lengths of these sequences can vary substantially, posing a computational challenge during training.

To illustrate this challenge, consider a scenario where we aim to fine-tune an LLM on a corpus of text. Each sentence in the corpus may have a different length, ranging from a few words to several dozen or even hundreds. Traditional approaches would require padding shorter sequences with special tokens up to the maximum sequence length. While this ensures uniformity for  processing, it introduces inefficiencies by including irrelevant padding tokens in the computation, which not only wastes  GPU resources but also can dilute the model's learning signal.

Batch-level, or dynamic, padding improves this by organizing sequences within a batch in a manner to improve computational efficiency without sacrificing learning efficacy. This is achieved by dynamically padding sequences to the same length within a batch, up to the maximum example length in each batch, allowing for parallel processing across multiple sequences. By eliminating the need for fixed-length padding across batches, Batch-level padding minimizes wasted computation on padding tokens, leading to improvements in training efficiency and batch inference throughput.

Central to the implementation of batch-level padding is the concept of masking. Masking mechanisms enable neural network models to selectively ignore padded regions during computation, ensuring that they do not contribute to the model's output or gradients. This enables the model to focus exclusively on the relevant parts of the input sequences, thereby preserving the integrity of the learning process while mitigating the effects of variable-length sequences. While batch-level padding is less wasteful than traditional padding, it is possible to improve computational efficiency further by packing small examples together, provided the masking mechanism is aware of the new example boundaries. 

Packing with Position IDs entails concatenating sequences into a single tensor and applying masking to appropriate position IDs to disregard  elements from other sequences during computation.  Packing improves upon batch-level padding,  accelerating the training process, allowing one to experiment with larger datasets and more complex models.
In summary,  Packing with Position IDs offers a pragmatic solution to the challenge of processing variable-length sequences efficiently and  unlocks higher levels of performance and scalability to LLM sequence processing. We provide a solution for this as well as a detailed analysis of its benefits.

%%%===========================
\section{Related Work}
%%%===========================

There are a few ways that sample packing can be enabled with Flash Attention 2 \cite{FA2} and proper attention masking of each example. The Flash Attention repository itself offers a way to pack while enabling proper masking of examples with Flash Attention. See Flash Attention Closed Issue 654 \cite{FAissue}. Others have proposed padding-free transformers, such as \cite{paddingfree,mayank}. The padding-free transformer methods require substantial and intrusive changes however to Hugging Face transformers library.
As such, these methods   have seen less uptake by the community and are not currently available in the  Hugging Face  library. 
For Hugging Face  Trainer users, it is desirable to have a readily available solution without requiring going outside the library. This feature is now available in Transformers version 4.44.

Most works on sample packing are concerned with how to select the sequences to pack together. This problem can be formulated as a  bin packing, or machine scheduling, problem from the combinatorial optimisation literature. 
For instance, \cite{packingWithoutCrossContamination2023ICLRreject} discuss one such approach for sequence selection and packing.  In \cite{scalingInstructionFinetunedLMs2024JMLR} the authors also use packing to combine multiple training examples into a single sequence, separating inputs from targets using an end-of-sequence token, and also using masking to prevent  tokens from attending to others across the packed example boundaries. 

The multi-pack sampler repository \cite{multipack}, used also by \cite{axolotl}, employs a first-fit-decreasing heuristic for the bin packing problem to select sequences to put together on each gpu in a distributed computing setting. Since this method is the most widely used, we make use  of it in conjunction with our Packing with PositionIDs solution to further enhance the packing performance.

A related effort for pre-training is studied by \cite{packpretrain} who note that using causal loss masking across different documents degrades performance of the models. The authors propose to limit causal masking to within documents during pre-training.
 We mention also the LengthGroupedSampler function \cite{grouplength} in Hugging Face Transformers library which is often used in conjunction with sample padding. We consider this method as another baseline, referred to as GroupByLength+Padding.

%%%================================================

\section{ Packing with Position IDs}

%%%================================================
We denote the input\_ids of the tokenized $i$-th example as a tensor of shape $(L_i,\,)$:
$$\X_i = [ \x_i^{(0)} , \cdots, \x_i^{(L_i-1)} ].$$
Let $\X_{i_1},\X_{i_2}, \cdots$ be an ordering of the training examples. Assuming a batch size  $bs=4$, consider a batch consisting of examples $\{\X_{i_1},\X_{i_2},\X_{i_3}, \X_{i_4}\}$. Then, in a batch-level padding based approach, the batch, $B$, is processed as a tensor of $(4, L_{max})$, where $L_{max} := max\{ L_{i_1}, L_{i_2}, L_{i_3}, L_{i_4}\}$ and the required number of padding tokens are appended to each example ($L_{max} - L_{i_1}$ to $\X_{i_1}$) to make them tensors with shape $(L_{max}, )$.

To avoid cross-contamination while packing examples, we propose to utilize position IDs to demarcate boundaries of individual examples in a packed sequence.
We assume support for Flash Attention and the availability of position IDs which is the case for the most popular open-source LLM models. 

Then, to  use the position IDs, we require arranging the data sequences accordingly. There are multiple ways to arrange the data with this solution, we discuss three below.

\subsection{Online minibatch Collating}
In minibatch collating, the padding-free collator must pack the examples online, for each minibatch, into a tensor of dimension, $\dim_{mini}$: 
$$\dim_{mini}=(1, \sum_{i=1\ldots 4} L_{i}),$$ 
and then provide position IDs which then take the form: 
$$[\,[\, 0, \, \cdots , \, (L_{i_1}-1),\,0, \, \cdots , \,(L_{i_2}-1), \, 0, \, \cdots , \, (L_{i_3}-1), \, 0, \, \cdots , \, (L_{i_4}-1) \,]\,].$$
This capability has been added to Hugging Face Trainer from Transformers 4.44, leveraging the new DataCollatorWithFlattening. Hugging Face users of the TRL library can also benefit from this capability by setting a new flag padding\_free=True in the DataCollatorForCompletionOnlyLM function. See Hugging Face blog \cite{HFblog}.

\subsection{Offline Batch Collating}
It is also  possible to pack a full set of samples offline in a single tensor of dimension $\dim_{flat}$: 
$$\dim_{flat}=(1, bs*msl),$$ 
where $bs$ is the batch size and $msl$ denotes the maximum sequence length allowed for individual training examples.

\subsection{Collating with Optimised Sample Selection}
One may wish to leverage  bin-packing-type sample selection algorithms such as \cite{multipack} in conjunction with the solution provided by Packing with PositionIDs. In this case the samples to assign to each gpu  and the position IDs for those samples are grouped per gpu. We call this Multipack+Position IDs.
To contrast with the Multipack bin packing method, we introduce two simple baselines: (q) \Randompack ~and (b) \Sortedpack; Examples are packed on a first-come-first-serve basis from randomly ordered training data set and sorted (length-wise from long to short) data set, respectively.

\begin{table}[ht]
\centering
\begin{tabular}{lcccc}
{Method} & {Attention Using} &  {Correct Cross-} & {No Broken } & {No Sorting}\\
{} & {Position IDs} & {Attention} &  {  Examples} & {of Dataset}\\
\hline \hline \\[-0.5pc] 
\Padding        & \xno  &  \xyes    & \xyes  & \xyes \\
\GroupByLength  & \xno  &  \xyes    & \xyes  & \xno \\
\PaddingFree    & \xyes &  \xyes    & \xyes  & \xyes \\ 
\\[-0.5pc]
\Packing        & \xno  &  \xno     & \xno   & \xyes \\
\PackingPosId   & \xyes &  \xyes    & \xno   & \xyes \\
\Multipack      & \xyes &  \xyes    & \xyes  & \xno  \\ 
\Sortedpack     & \xyes &  \xyes    & \xyes  & \xno  \\
\Randompack     & \xyes &  \xyes    & \xyes  & \xyes \\
\\[-0.5pc] \hline
\end{tabular}
\vspace{3mm}
\caption{Characteristics of different methods of grouping examples into batches}
\label{tab:packing_methods}
\end{table}

Specifically, to enable the use of position IDs, we  modify the models’ \_flash\_attention\_forward(), 
adding the argument position\_ids and  
extracting the number of examples in the batch from the position\_ids. When attention\_mask is None
in the case of number of examples > batch size, we compute cu\_seq\_len from position\_ids and use the flash\_attn\_varlen\_func().

\subsection{PaddingFreeCollator}

 We   provide  a new off-the-shelf data collator, the PaddingFreeCollator,  summarised below.

\begin{center}
~~~ return \{ \emph{input\_ids}: ..., ~\emph{labels}: ..., ~\emph{position\_ids}: ... \}

\emph{input\_ids} (1, $\sum$ sequence\_length\_of\_each\_example) : \\
Concatenate all the examples in the mini.

\emph{labels} (1, $\sum$ sequence\_length\_of\_each\_examples) : \\
Convert the first label\_id of each example into -100, then concatenate.

\emph{position\_ids} (1, $\sum$ sequence\_length\_of\_each\_examples) : \\
Generate position ids for each example and concatenate them all.
\end{center}

Next we illustrate the simplicity of the method with an example, below.

\subsection{Example}
Consider again the  scenario with a batchsize = 4 where the  sequences are as follows:\\
$$[[ 10, 11, 12, 13 ],$$
$$[ 20, 21, 22, 23, 24, 25, 26, 27 ],$$ 
$$[ 30, 31, 32, 33, 34 ],$$
$$  [ 40, 41, 42, 43, 44, 45, 46, 47, 48, 49, 410 ]]$$
The padding-free collator returns the input IDs, labels, and the position IDs of each example after concatenating the examples together. Hence, the collator provides:\\
\emph{input\_ids}: $$[[ 10, 11, 12, 13, 20, 21, 22, 23, 24, 25, 26,27, $$
$$ 30, 31, 32, 33, 34, 40, 41, 42, 43, 44, 45, 46, 47, 48, 49, 410 ]],$$ \\
\emph{labels}: $$[[ -100, 11, 12, 13, -100, 21, 22, 23, 24, 25, 26, 27, $$ $$-100, 31, 32, 33, 34, -100, 41, 42, 43, 44, 45, 46, 47, 48, 49, 410 ]],$$ \\
\emph{position\_ids}: $$[[ 0, 1, 2, 3, 0, 1, 2, 3, 4, 5, 6, 7, $$ 
$$0, 1, 2, 3, 4, 0, 1, 2, 3, 4, 5, 6, 7, 8, 9, 10 ]].$$

The modifications required are lightweight and are limited to providing the position IDs to Flash Attention. For details,  see  the repository at 
\cite{RDrepo}. This solution relies, however, on the model exposing position IDs. As of the time of writing, 14 models expose position IDs and are supported by the solution. Specifically:  llama, mistral, mixtral, granite, dbrx, falcon, gemma, olmo, phi, phi2, phi3, qwen2, qwen2\_moe, stablelm, starcoder2 are all supported by the solution.

%%%================================================
\section{ Experiments and Results}
%%%================================================

We consider three different datasets with different example-length characteristics: FLAN \cite{flan}, OrcaMath\cite{orca}, and  python code data from The Stack\cite{bigcodestack}. 
 A subset of 20K examples from each of the 3 datasets was used in the experiments. We evaluate the throughput in tokens/second, denoted Tok/s, peak memory, denoted Mem, and validation loss, denoted VLoss. All tests were performed on a single A100-80GB node having 8 GPU with FSDP. The throughput measurements are averaged over the 8 GPU and the peak memory recorded is the maximum over the 8 GPU.

\begin{figure}[ht]
\centering
\begin{minipage}[b]{.5\linewidth}
\centering
\includegraphics[width=\linewidth]{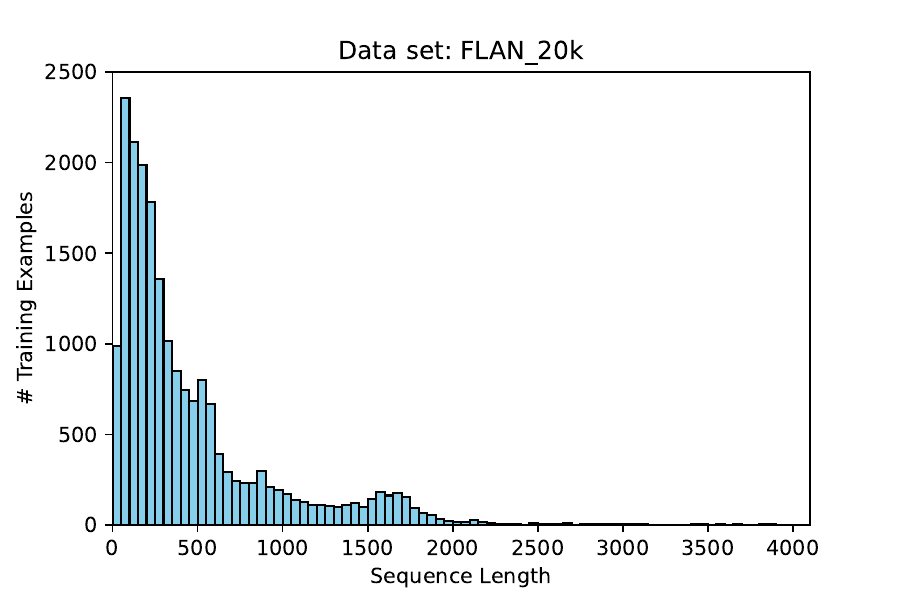}
\end{minipage}\hfill
\begin{minipage}[b]{.5\linewidth}
\centering
\includegraphics[width=\linewidth]{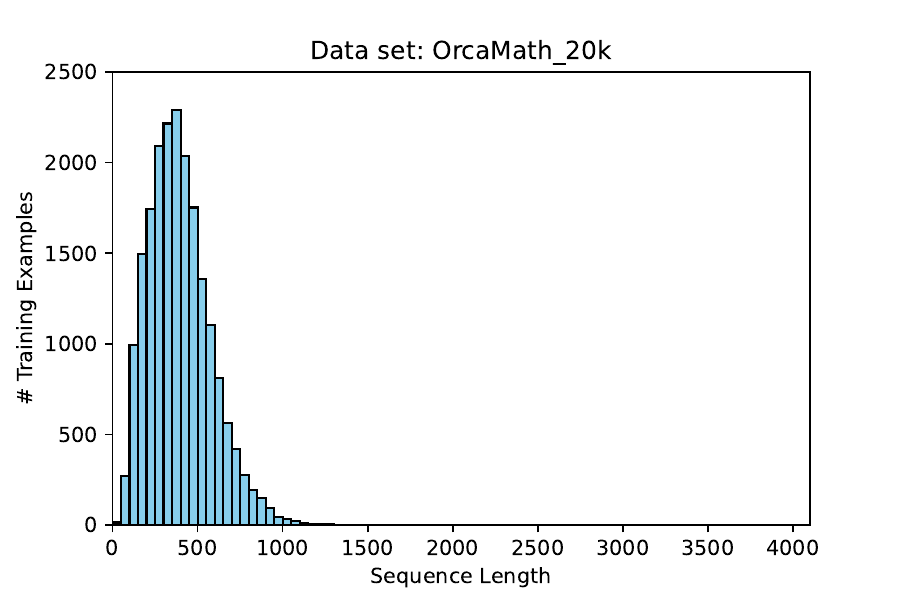}
\end{minipage}
\centering
\begin{minipage}[b]{.5\linewidth}
\centering
\includegraphics[width=\linewidth]{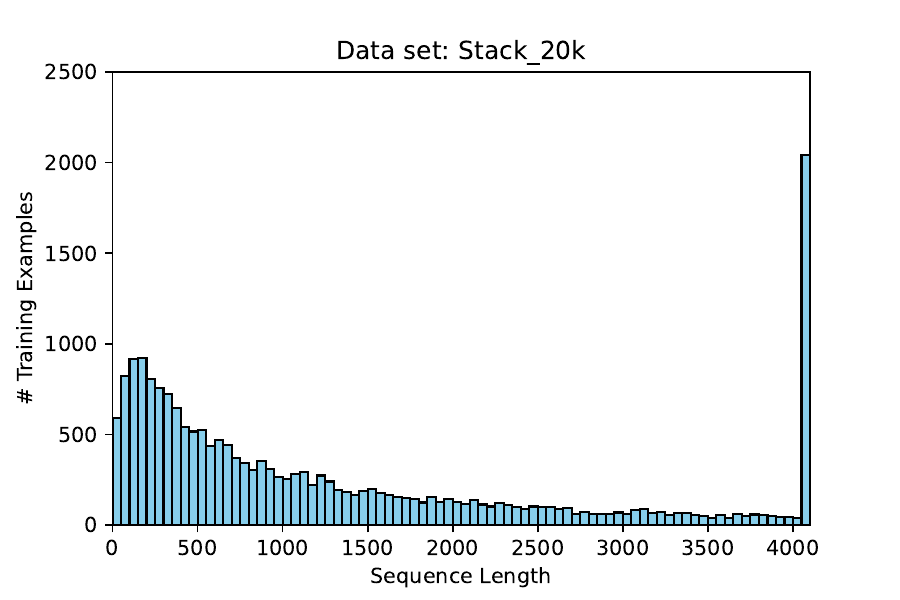}
\end{minipage}\hfill
\caption{Histograms of sequence lengths for the 3 training datasets: (top left) FLAN\_20k, (top right) OrcaMath\_20k, and (bottom) the Stack\_20k. }
\label{fig:dataset_histogram}
\end{figure}

The characteristics of example lengths in our 20K subsets of FLAN, OrcaMath, as well as in the python coding dataset from The Stack are shown in the histograms in Figure \ref{fig:dataset_histogram}. Observe that both FLAN and OrcaMath have primarily small example lengths,  which makes them amenable to significant throughput gains via packing, while the coding dataset the Stack has significantly longer examples, and hence we expect less throughput benefit from packing on that dataset.

%%%%%%%%%
\subsection{Main Results}
%%%%%

Four approaches were evaluated. The two base cases are: (i) No packing with truncate, i.e., padding, and (ii) Basic packing (as is currently available in SFT trainer, without position IDs).
Then, we evaluate two variants of the proposed solution: (iii) Our solution providing packing with position IDs, in which the batch is prepared offline and flattened, and (iv) Our solution with online packing with position IDs for each minibatch. 

We expect that base case (i) using padding will be the slowest approach but provide optimal training loss reduction since it does not disrupt any training examples. On the other hand, basic packing without position IDs can achieve high throughput, at the expense of distorting examples through improper attention masks.

We demonstrate the benefits of the solution  on 10 of the 14 supported models. Specifically, we test it on \texttt{ Llama-2-7B-fp16, mistralai/Mistral-7B-v0.1, 
granite-8b-code-base, tiiuae/falcon-7b, google/gemma-7b, microsoft/phi-2, Qwen/CodeQwen1.5-7B, bigcode/starcoder2-7b, stabilityai/stablelm-2-1\_6b,  Qwen/Qwen1.5-MoE-A2.7B}.

Table \ref{tab:model_results_flan} provides the main results on a 20K subset of the FLAN instruct tuning dataset \cite{flan} and Table \ref{tab:model_results_orca} shows the analogous results on the OrcaMath dataset \cite{orca}. For both datasets, all models are run for one epoch with gradient accumulation steps ({\bf gas}) of 2, maximum sequence length ({\bf msl}) of 4096 and minibatch size ({\bf bs}) per GPU set to 4.

Observe that the Packing with PositionID methods improve the throughput substantially  in terms of tokens per second, in many case above and beyond even basic packing that does not account for improper cross-attention (the second row in each model section). 

However, this maximal packing has an impact on the loss behaviour. Due to the fact that far fewer optimisation steps are taken with such maximal packing, the loss does not decrease as fast, and its effect is confirmed by the validation loss ("VLoss") after one epoch.  Hence, as a best-of-both worlds remedy, we propose the online minibatch approach to Packing with PositionIDs. Due to the packing of fewer examples in each pack, minibatch packing does not achieve the maximal throughput, however it achieves the same optimal loss pattern and hence validation loss, as the inefficient padding-based approach. It also improves on throughput substantially beyond the padding-based approach.

The variation in benefit from our proposed approach across the FLAN and OrcaMath datasets can be attributed to the statistical dataset characteristics of each; the very small sample lengths of FLAN, whose primary mode is around 100 tokens, is highly amenable to significant throughput increase from packing, more so than OrcaMath, whose mode is around 400 tokens, as seen in Figure \ref{fig:dataset_histogram}. This means that offline packing can achieve very significant throughput improvements. On the other hand, with minibatch packing, the padding baseline pads only to the length of the longest sequence in each minibatch. Hence, the variance (rather than the mean) of the sample lengths affects the gain in throughput, with high variance leading to more minibatch padding. This is also confirmed by the higher improvement in throughput from minibatch packing with PositionIDs compared to padding in FLAN, Table \ref{tab:model_results_flan} versus OrcaMath, in Table \ref{tab:model_results_orca}.
The benefits across model architectures are  consistent, with the exception of Gemma-7B and Qwen1.5-MoE-A2.7B.

\begin{table}[h]
\begin{tabular}{  l | l  l  l | r r r r r}
	Model & Pack? & inp\_ids & pos\_id? & Rows & Time & Tok/s & Mem & VLoss \\ \hline \hline
 
	llama2 & no & bs, $\max(B)$ & no & 19958 & 1526 & 771 & 29234 & 1.266 \\
	 & yes & bs, seq & no & 2346 & 388 & 3091 & 29433 & 1.579 \\ 
	 & flat & 1, bs*msl & 1, bs*msl & 599 & 366 & 3222 & 29438 & 1.578\\ 
	 & mini & 1, $\sum(B)$ & 1, $\sum(B)$ & 19958 & 809 & 1455 & 23854 & 1.262 \\ \hline
  
	mistral & no & bs, $\max(B)$ & no & 19961 & 1553 & 742 & 30625 & 1.129 \\ 
	 & yes & bs, seq & no & 2294 & 392 & 2986 & 30783 & 1.306 \\ 
	 & flat & 1, bs*msl & 1, bs*msl & 585 & 383 & 3010 & 30783 & 1.284 \\ 
	 & mini & 1, $\sum(B)$ & 1, $\sum(B)$ & 19961 & 818 & 1408 & 24549 & 1.127 \\ \hline

  granite & no & bs, $\max(B)$ & no & 19947 & 1741 &	693&	36277	& 1.169 \\ 
	 & yes & bs, seq & no & 
  2420&	471	&2628	&36278	&1.740
   \\ 
	 & flat & 1, bs*msl & 1, bs*msl & 
  620&	459	&2657	&36276	&1.767
   \\ 
	 & mini & 1, $\sum(B)$ & 1, $\sum(B)$ &
  19947	&888	&1358	&27538&	1.169
  \\ \hline
	falcon & no & bs, $\max(B)$ & no & 19969 & 1229 & 890 & 29894 & 1.842\\ 
	 & yes & bs, seq & no & 2168 & 326 & 3401 & 30106 & 2.585 \\ 
	 & flat & 1, bs*msl & 1, bs*msl & 536 & 293 & 3639 & 30202 & 2.415 \\ 
	 & mini & 1, $\sum(B)$ & 1, $\sum(B)$ & 19969 & 687 & 1592 & 24622 & 1.841 \\ \hline
  
	gemma & no & bs, $\max(B)$ & no & 19974 & 1671 & 602 & 49982 & 1.930 \\ 
	 & yes & bs, seq & no & 1990 & 461 & 2211 & 50014 & 3.464 \\ 
	 & flat & 1, bs*msl & 1, bs*msl & 1056 & 554 & 1824 & 50020 & 3.418 \\ 
	 & mini & 1, $\sum(B)$ & 1, $\sum(B)$ & 19974 & 1223 & 823 & 41628 & 1.923 \\ \hline
  
	phi2 & no & bs, $\max(B)$ & no & 19974 & 741 & 1449 & 20228 & 2.948 \\ 
	 & yes & bs, seq & no & 2123 & 187 & 5806 & 20293 & 3.375 \\ 
	 & flat & 1, bs*msl & 1, bs*msl & 525 & 172 & 6066 & 20342 & 3.464\\ 
	 & mini & 1, $\sum(B)$ & 1, $\sum(B)$ & 19974 & 415 & 2588 & 13315 & 2.940\\ \hline
  
	CodeQwen & no & bs, $\max(B)$ & no & 19966 & 1477 & 732 & 44549 & 1.578 \\ 
	 & yes & bs, seq & no & 2151 & 368 & 2992 & 44575 & 1.895 \\ 
	 & flat & 1, bs*msl & 1, bs*msl & 544 & 367 & 2942 & 44579 & 1.855 \\ 
	 & mini & 1, $\sum(B)$ & 1, $\sum(B)$ & 19966 & 784 & 1380 & 31652 & 1.577 \\ \hline
  
	starcoder & no & bs, $\max(B)$ & no & 19947 & 1585 & 761 & 34122 & 2.085 \\ 
	 & yes & bs, seq & no & 2420 & 417 & 2965 & 34122 & 2.547 \\ 
	 & flat & 1, bs*msl & 1, bs*msl & 620 & 396 & 3072 & 34121 & 2.483\\ 
	 & mini & 1, $\sum(B)$ & 1, $\sum(B)$ & 19947 & 835 & 1445 & 25484 & 2.090  \\ \hline
  
	stablelm& no & bs, $\max(B)$ & no & 19969 & 369 & 2820 & 16488 & 1.064\\ 
	 & yes & bs, seq & no & 2064 & 91 & 11595 & 16682 & 1.252 \\ 
	 & flat & 1, bs*msl & 1, bs*msl & 522 & 85 & 12100 & 16700 & 1.201 \\ 
	 & mini & 1, $\sum(B)$ & 1, $\sum(B)$ & 19969 & 229 & 4540 & 11011 & 1.066 \\ \hline
  
	QwenMOE & no & bs, $\max(B)$ & no & 19969 & 4066 & 254 & 55436 & 1.268\\ 
	 & yes & bs, seq & no & 2045 & 497 & 2101 & 55733 & 1.592 \\ 
	 & flat & 1, bs*msl & 1, bs*msl & 1083 & 607 & 1706 & 55691 & 1.556 \\ 
	 & mini & 1, $\sum(B)$ & 1, $\sum(B)$ & 19969 & 4014 & 257 & 50774 & 1.268 \\ \hline
\end{tabular}
\vspace{3mm}
\caption{Summary of main results on a 20K sample of the FLAN dataset over 10 of the 14 supported models. The table  compares padding each example (first row, each  section) with basic packing without position IDs (second row, each section), and two variants of the solution with position IDs: offline flattening and the online minibatch. }
\label{tab:model_results_flan}
\end{table}

\begin{table}[h]
\begin{tabular}{  l | l  l  l | r r r r r}
	Model & Pack? & inp\_ids & pos\_id? & Rows & Time & Tok/s & Mem & VLoss \\ \hline \hline
 
	llama2 & no & bs, $\max(B)$ & no & 20000 & 790 & 1269 & 22305 & 0.458\\
	 & yes & bs, msl & no & 1946 & 315 & 3155 & 29433 & 0.550\\
	 &flat & 1, bs*msl & 1, bs*msl & 488 & 295 & 3324 & 29432 & 0.521\\
	 & mini & 1, $\sum(B)$ & 1, $\sum(B)$ & 20000 & 574 & 1746 & 20950 & 0.459\\ \hline
  
	mistral & no & bs, $\max(B)$ & no & 20000 & 812 & 1216 & 23603 & 0.321\\
	 & yes & bs, msl & no & 1921 & 326 & 3006 & 30783 & 0.353\\
	 &flat & 1, bs*msl & 1, bs*msl & 478 & 307 & 3137 & 30783 & 0.350 \\
	 & mini & 1, $\sum(B)$ & 1, $\sum(B)$ & 20000 & 596 & 1658 & 22409 & 0.320 \\ \hline

  granite & no & bs, $\max(B)$ & no &
  20000&	809&	1199&	26489	&0.3242
 \\ 
	 & yes & bs, seq & no & 
  1887&	369	&2614	&36278	&0.3928
   \\ 
	 & flat & 1, bs*msl & 1, bs*msl & 
  473&	351&	2712	&36277	&0.383
 \\ 
	 & mini & 1, $\sum(B)$ & 1, $\sum(B)$ &
  20000	&575&	1689	&24423&	0.3242
   \\ \hline
  
	falcon & no & bs, $\max(B)$ & no & 20000 & 661 & 1428 & 22770 & 0.618\\
	 & yes & bs, msl & no & 1834 & 279 & 3358 & 30106 & 0.744\\
	 &flat & 1, bs*msl & 1, bs*msl & 456 & 248 & 3700 & 30201 & 0.695\\
	 & mini & 1, $\sum(B)$ & 1, $\sum(B)$ & 20000 & 500 & 1887 & 21702 & 0.618 \\ \hline
  
	gemma & no & bs, $\max(B)$ & no & 20000 & 1076 & 858 & 32325 & 0.344\\
	 & yes & bs, msl & no & 1794 & 420 & 2188 & 50014 & 0.962\\
	 &flat & 1, bs*msl & 1, bs*msl & 919 & 470 & 1944 & 50020 & 0.798\\
	 & mini & 1, $\sum(B)$ & 1, $\sum(B)$ & 20000 & 949 & 972 & 30794 & 0.345 \\ \hline
  
	phi2 & no & bs, $\max(B)$ & no & 20000 & 371 & 2240 & 11209 & 0.479\\
	 & yes & bs, msl & no & 1614 & 143 & 5755 & 20293 & 0.561\\
	 &flat & 1, bs*msl & 1, bs*msl & 399 & 129 & 6225 & 20341 & 0.514 \\ \
	 & mini & 1, $\sum(B)$ & 1, $\sum(B)$ & 20000 & 306 & 2714 & 9817 & 0.483 \\ \hline
  
	CodeQwen & no & bs, $\max(B)$ & no & 20000 & 750 & 1187 & 28760 & 0.483\\
	 & yes & bs, msl & no & 1730 & 295 & 3001 & 44575 & 0.541\\
	 &flat & 1, bs*msl & 1, bs*msl & 432 & 288 & 3028 & 44579 & 0.522 \\ 
	 & mini & 1, $\sum(B)$ & 1, $\sum(B)$ & 20000 & 559 & 1594 & 26132 & 0.483 \\ \hline
  
	starcoder & no & bs, $\max(B)$ & no & 20000 & 782 & 1271 & 24475 & 0.504\\
	 & yes & bs, msl & no & 1933 & 330 & 2988 & 34122 & 0.563\\
	 &flat & 1, bs*msl & 1, bs*msl & 483 & 302 & 3219 & 34123 & 0.558\\
	 & mini & 1, $\sum(B)$ & 1, $\sum(B)$ & 20000 & 575 & 1729 & 22439 & 0.504 \\ \hline
  
	stablelm & no & bs, $\max(B)$ & no & 20000 & 226 & 3970 & 8986 & 0.534 \\ 
	 & yes & bs, msl & no & 1744 & 77 & 11578 & 16682 & 0.573\\
	 &flat & 1, bs*msl & 1, bs*msl & 434 & 70 & 12470 & 16700 & 0.567\\
	 & mini & 1, $\sum(B)$ & 1, $\sum(B)$ & 20000 & 195 & 4614 & 7368 & 0.534 \\ \hline
  
	QwenMOE & no & bs, $\max(B)$ & no & 20000 & 4165 & 216 & 45152 & 0.335\\
	 & yes & bs, msl & no & 1744 & 444 & 2005 & 55733 & 0.366\\
	 &flat & 1, bs*msl & 1, bs*msl & 889 & 484 & 1826 & 55692 & 0.352\\
	 & mini & 1, $\sum(B)$ & 1, $\sum(B)$ & 20000 & 3849 & 234 & 44572 & 0.335 \\ \hline
\end{tabular}
\vspace{3mm}
\caption{Summary of main results on a 20K sample of the OrcaMath dataset over 10 of the 14 supported models. The table compares padding each example (first row, each  section) with basic packing without position IDs (second row, each section), and two variants of the solution with position IDs: offline flattening and the online minibatch. }
\label{tab:model_results_orca}
\end{table}

%%----------------------------------------------------
\subsection{Ablation Study on Minibatch Size}
\begin{figure}[ht]
\centering
\includegraphics[width=0.5\linewidth]{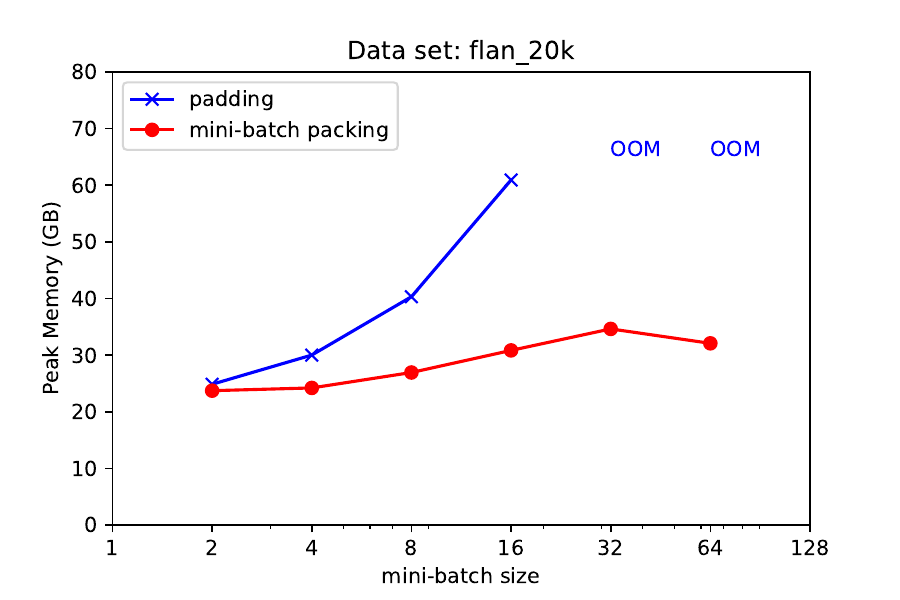}
\caption{Minibatch packing offers lower GPU peak memory usage than padding approach as minibatch size increases. ``OOM" denotes out of memory error. Lower is better.}
\label{fig:mem_vs_bs}
\end{figure}

\begin{figure}[ht]
\centering
\begin{minipage}[b]{.5\linewidth}
\centering
\includegraphics[width=\linewidth]{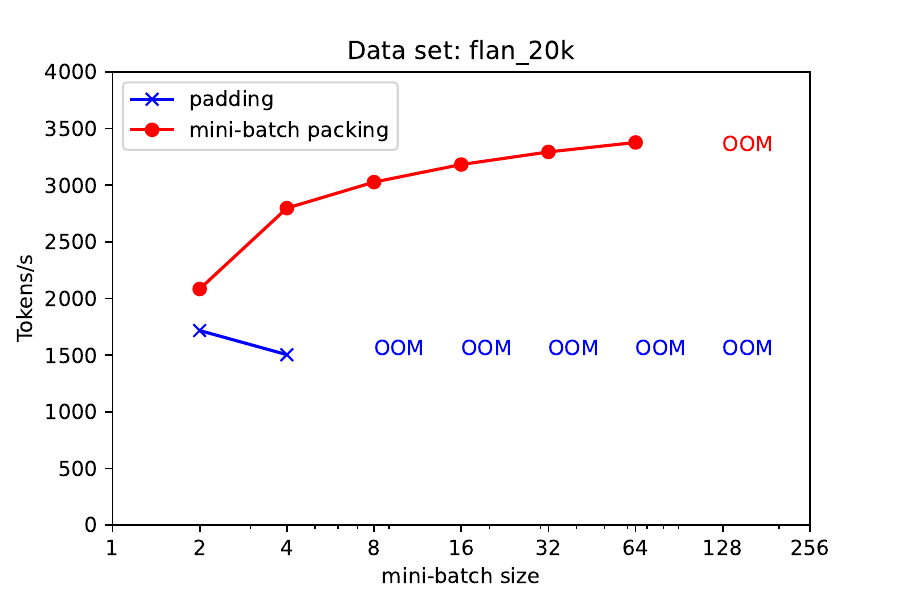}
\end{minipage}\hfill
\begin{minipage}[b]{.5\linewidth}
\centering
\includegraphics[width=\linewidth]{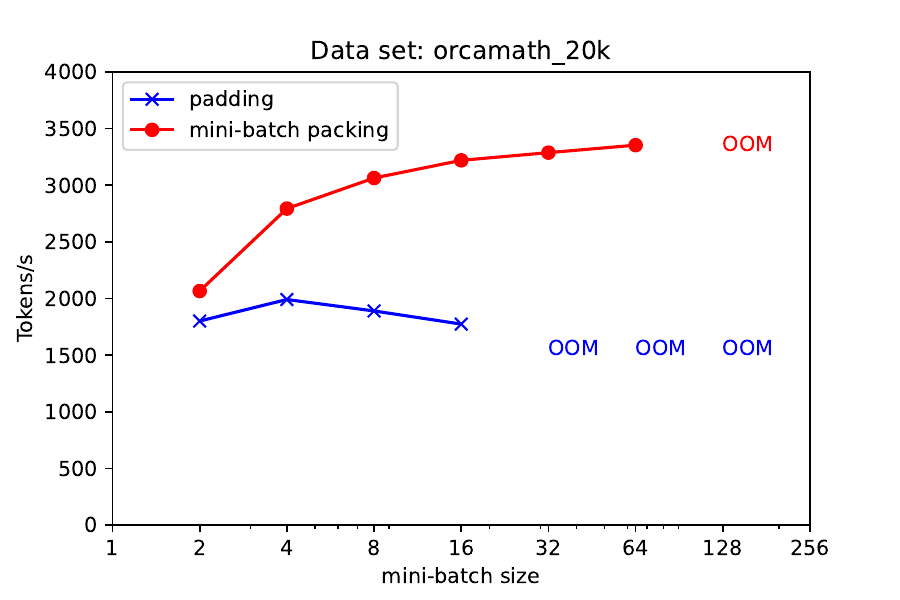}
\end{minipage}
\caption{Minibatch packing offers better throughput (Tokens/s) than padding  as minibatch size increases. ``OOM" denotes out of memory error. Higher is better.}
\label{fig:throughput_vs_bs}
\end{figure}

%%------------------------------------------------
In this ablation study we compare the effect of minibatch size (bs) on training throughput and memory for two approaches: the standard padding vs proposed minibatch packing. For this experiment, we fine-tune  Mistral-7B\cite{mistral}  on 2 different training datasets: FLAN\_20k, (right) OrcaMath\_20k. We present the throughput results in Figure~\ref{fig:throughput_vs_bs} and the memory results in \ref{fig:mem_vs_bs}. The advantage of  minibatch packing is significant, not only is the throughput a factor of 2x faster and the peak memory usage significantly lower than using padding, but the training runs out of memory (OOM) much later allowing for larger batch sizes.

%%----------------------------------------------------
\subsection{Ablation Study on Sample Selection Method}
%%------------------------------------------------
In this ablation study we consider two different models: Mistral-7B\cite{mistral} and CodeLlama-Python-7B  \cite{rozière2023code}.
In Tables \ref{tab:flan_20k},\ref{tab:orcamath_20k}, and \ref{tab:stack_20k},  we examine different batch sizes and sequence lengths on two of the most popular models, Mistral-7B and CodeLlama-Python-7B with several sample selection methods. We provide the results on FLAN, OrcaMath, and the Stack \cite{bigcodestack}, respectively. In the tables, {\bf bs} is the minibatch size per GPU, {\bf msl} stands for maximum sequence length of an example/packed sequence, and {\bf gas} denotes the number gradient accumulation steps. Note that the Stack dataset has   high variance in sequence length as seen in Figure \ref{fig:dataset_histogram},  and   long sequence lengths.

As expected, the padding-based approaches have the lowest training throughput, and offline packing offers the highest throughput. Also as expected and as seen previously minibatch packing offers an intermediate ground with still-signicant throughput increases.  

It is instructive to delve into both training and validation loss. The offline packing methods, even using position IDs fail to provide the same level of validation loss after one epoch that padding and minibatch packing with position IDs provide. As noted above, this is due primarily to the fewer number of optimisation update steps performed. 

Finally, we examine how much  benefit can be achieved by coupling the use of Packing with PositionIDs with the bin-packing-based sample selection of Multi-pack sampler \cite{multipack}. The first observation is that the benefit, across all three datasets, is rather limited. However, while the throughput decreases very slightly when using the Multi-pack sampler, there is on average a small benefit to the loss behaviour of the models using bin-packing-based sampling.

\begin{table}[ht]
\centering
\begin{tabular}{l c | c c c c c}

Method         & (bs, msl, gas) & Steps & Time  & Tok/s & Train Loss & V Loss \\ \hline \hline

\Padding       & (2, 4096, 32)  & 311   & 5516  & 1671  & 1.271      & 1.117  \\
\GroupByLength & (2, 4096, 32)  & 311   & 3997  & 2306  & 1.277      & 1.120  \\ 
\PaddingFree   & (2, 4096, 32)  & 311   & 4028  & 2289  & 1.273      & 1.123  \\ \hline
\Packing       & (2, 4096, 32)  & 35    & 2895  & 3184  & 1.372      & 1.294  \\
\Packing       & (1, 8192, 32)  & 35    & 3058  & 3014  & 1.362      & 1.352  \\
\PackingPosId  & (2, 4096, 32)  & 35    & 2888  & 3191  & 1.378      & 1.221  \\
\PackingPosId  & (1, 8192, 32)  & 35    & 2781  & 3314  & 1.363      & 1.221  \\
\Multipack     & (1, 8192, 32)  & 37    & 2782  & 3314  & 1.335      & 1.215  \\
\Sortedpack    & (1, 8192, 32)  & 37    & 2842  & 3243  & 1.318      & 1.278  \\
\Randompack    & (1, 8192, 32)  & 37    & 2852  & 3233  & 1.335      & 1.208  \\ \hline
\Packing       & (2, 4096, 4)   & 281   & 2959  & 3115  & 1.339      & 1.252  \\
\Packing       & (1, 8192, 4)   & 281   & 3143  & 2932  & 1.331      & 1.288  \\
\PackingPosId  & (2, 4096, 4)   & 281   & 2845  & 3239  & 1.338      & 1.170  \\
\PackingPosId  & (1, 8192, 4)   & 281   & 2839  & 3246  & 1.321      & 1.154  \\
\Multipack     & (1, 8192, 4)   & 299   & 2953  & 3122  & 1.294      & 1.153  \\
\Sortedpack    & (1, 8192, 4)   & 300   & 2938  & 3138  & 1.264      & 1.210  \\
\Randompack    & (1, 8192, 4)   & 300   & 2912  & 3166  & 1.293      & 1.151  \\ \hline \hline

\Padding       & (4, 4096, 16)  & 311   & 6184  & 1491  & 1.295      & 1.127  \\
\GroupByLength & (4, 4096, 16)  & 311   & 3394  & 2716  & 1.315      & 1.131  \\
\PaddingFree   & (4, 4096, 16)  & 311   & 3310  & 2785  & 1.296      & 1.127  \\ \hline
\Packing       & (4, 4096, 16)  & 35    & 2886  & 3193  & 1.369      & 1.307  \\ 
\Packing       & (1, 16384, 16) & 35    & 3042  & 3027  & 1.358      & 1.431  \\
\PackingPosId  & (4, 4096, 16)  & 35    & 2699  & 3415  & 1.375      & 1.225  \\
\PackingPosId  & (1, 16384, 16) & 35    & 2683  & 3432  & 1.350      & 1.225  \\
\Multipack     & (1, 16384, 16) & 36    & 2700  & 3414  & 1.337      & 1.217  \\
\Sortedpack    & (1, 16384, 16) & 36    & 2688  & 3429  & 1.312      & 1.281  \\
\Randompack    & (1, 16384, 16) & 36    & 2722  & 3386  & 1.332      & 1.223  \\ \hline
\Packing       & (4, 4096, 2)   & 281   & 2878  & 3202  & 1.340      & 1.255  \\
\Packing       & (1, 16384, 2)  & 281   & 3109  & 2961  & 1.324      & 1.363  \\
\PackingPosId  & (4, 4096, 2)   & 281   & 2759  & 3340  & 1.340      & 1.169  \\
\PackingPosId  & (1, 16384, 2)  & 281   & 2833  & 3250  & 1.313      & 1.145  \\
\Multipack     & (1, 16384, 2)  & 290   & 2772  & 3326  & 1.299      & 1.154  \\
\Sortedpack    & (1, 16384, 2)  & 290   & 2822  & 3267  & 1.268      & 1.193  \\
\Randompack    & (1, 16384, 2)  & 290   & 2794  & 3299  & 1.294      & 1.156  \\ \hline

\end{tabular}
\vspace{3mm}
\caption{Finetuning Mistral-7B for 1 epoch on FLAN\_20k dataset: performance comparison of different sample packing approaches. The first group in each section uses minibatches and the second and third groups use offline packing.  Throughput   (train tokens/second) is improved substantially using  Packing  with Position IDs. Loss performance is the same for padding and  minibatch packing.}
\label{tab:flan_20k}
\end{table}

\begin{table}[ht]
\centering
\begin{tabular}{l c | c c c c c}

Method         & (bs, msl, gas) & Steps & Time  & Tok/s & Train Loss & V Loss \\ \hline \hline 

\Padding       & (2, 4096, 32)  & 312   & 4195  & 1884  & 0.331      & 0.320  \\
\GroupByLength & (2, 4096, 32)  & 312   & 3680  & 2148  & 0.335      & 0.326  \\
\PaddingFree   & (2, 4096, 32)  & 312   & 3683  & 2146  & 0.331      & 0.320  \\ \hline
\Packing       & (2, 4096, 32)  & 30    & 2525  & 3130  & 0.383      & 0.352  \\
\Packing       & (1, 8192, 32)  & 30    & 2649  & 2985  & 0.369      & 0.359  \\
\PackingPosId  & (2, 4096, 32)  & 30    & 2359  & 3351  & 0.396      & 0.348  \\
\PackingPosId  & (1, 8192, 32)  & 30    & 2341  & 3378  & 0.383      & 0.348  \\
\Multipack     & (1, 8192, 32)  & 31    & 2397  & 3297  & 0.369      & 0.347  \\
\Sortedpack    & (1, 8192, 32)  & 31    & 2446  & 3232  & 0.370      & 0.347  \\
\Randompack    & (1, 8192, 32)  & 31    & 2401  & 3292  & 0.368      & 0.346  \\ \hline
\Packing       & (2, 4096, 4)   & 241   & 2552  & 3098  & 0.355      & 0.331  \\
\Packing       & (1, 8192, 4)   & 241   & 2676  & 2954  & 0.343      & 0.333  \\
\PackingPosId  & (2, 4096, 4)   & 241   & 2443  & 3236  & 0.356      & 0.323  \\
\PackingPosId  & (1, 8192, 4)   & 241   & 2494  & 3169  & 0.344      & 0.322  \\
\Multipack     & (1, 8192, 4)   & 248   & 2608  & 3031  & 0.331      & 0.323  \\
\Sortedpack    & (1, 8192, 4)   & 248   & 2460  & 3214  & 0.332      & 0.323  \\
\Randompack    & (1, 8192, 4)   & 248   & 2471  & 3199  & 0.330      & 0.323  \\ \hline \hline

\Padding       & (4, 4096, 16)  & 312   & 4048  & 1953  & 0.330      & 0.320  \\
\GroupByLength & (4, 4096, 16)  & 312   & 2929  & 2699  & 0.335      & 0.325  \\
\PaddingFree   & (4, 4096, 16)  & 312   & 3142  & 2516  & 0.330      & 0.320  \\ \hline
\Packing       & (4, 4096, 16)  & 30    & 2500  & 3163  & 0.383      & 0.352  \\
\Packing       & (1, 16384, 16) & 30    & 2682  & 2945  & 0.363      & 0.374  \\
\PackingPosId  & (4, 4096, 16)  & 30    & 2365  & 3343  & 0.396      & 0.348  \\
\PackingPosId  & (1, 16384, 16) & 30    & 2358  & 3349  & 0.376      & 0.347  \\
\Multipack     & (1, 16384, 16) & 30    & 2264  & 3492  & 0.370      & 0.347  \\
\Sortedpack    & (1, 16384, 16) & 30    & 2253  & 3509  & 0.371      & 0.348  \\
\Randompack    & (1, 16384, 16) & 30    & 2240  & 3529  & 0.369      & 0.347  \\ \hline
\Packing       & (4, 4096, 2)   & 241   & 2559  & 3089  & 0.355      & 0.330  \\
\Packing       & (1, 16384, 2)  & 241   & 2697  & 2929  & 0.337      & 0.346  \\
\PackingPosId  & (4, 4096, 2)   & 241   & 2362  & 3346  & 0.356      & 0.323  \\
\PackingPosId  & (1, 16384, 2)  & 241   & 2331  & 3388  & 0.337      & 0.322  \\
\Multipack     & (1, 16384, 2)  & 245   & 2469  & 3202  & 0.330      & 0.323  \\
\Sortedpack    & (1, 16384, 2)  & 245   & 2379  & 3323  & 0.333      & 0.327  \\
\Randompack    & (1, 16384, 2)  & 245   & 2340  & 3378  & 0.330      & 0.322  \\ \hline 

\end{tabular}
\vspace{3mm}
\caption{Finetuning Mistral-7B for 1 epoch on OrcaMath\_20k dataset: performance comparison of different sample packing approaches. The first group in each section uses minibatches and the second and third groups use offline packing.  Throughput   (train tokens/second) is improved substantially using  Packing  with Position IDs. Loss performance is the same for padding and  minibatch packing.}
\label{tab:orcamath_20k}
\end{table}

\begin{table}[ht]
\centering
\begin{tabular}{l c | c c c c c}
Method         & (bs, msl, gas) & Steps & Time  & Tok/s & Train Loss & V Loss \\ \hline \hline

\Padding       & (2, 4096, 32)  & 312   & 12463 & 2167  & 0.440      & 0.505  \\
\GroupByLength & (2, 4096, 32)  & 312   & 8886  & 3040  & 0.535      & 0.505  \\
\PaddingFree   & (2, 4096, 32)  & 312   & 8845  & 3054  & 0.440      & 0.505  \\ \hline
\Packing       & (2, 4096, 32)  & 103   & 8157  & 3311  & 0.441      & 0.581  \\
\Packing       & (1, 8192, 32)  & 103   & 9117  & 2962  & 0.420      & 0.588  \\
\PackingPosId  & (2, 4096, 32)  & 103   & 8157  & 3311  & 0.420      & 0.588  \\
\PackingPosId  & (1, 8192, 32)  & 103   & 8129  & 3322  & 0.414      & 0.541  \\
\Multipack     & (1, 8192, 32)  & 121   & 8227  & 3283  & 0.393      & 0.539  \\
\Sortedpack    & (1, 8192, 32)  & 112   & 8224  & 3285  & 0.393      & 0.541  \\
\Randompack    & (1, 8192, 32)  & 122   & 8500  & 3178  & 0.392      & 0.538  \\ \hline
\Packing       & (2, 4096, 12)  & 274   & 8170  & 3306  & 0.436      & 0.574  \\
\Packing       & (1, 8192, 12)  & 274   & 8885  & 3040  & 0.415      & 0.583  \\
\PackingPosId  & (2, 4096, 12)  & 274   & 8053  & 3354  & 0.432      & 0.516  \\
\PackingPosId  & (1, 8192, 12)  & 274   & 8058  & 3352  & 0.411      & 0.515  \\
\Multipack     & (1, 8192, 12)  & 324   & 8617  & 3135  & 0.389      & 0.514  \\
\Sortedpack    & (1, 8192, 12)  & 299   & 8293  & 3257  & 0.389      & 0.518  \\
\Randompack    & (1, 8192, 12)  & 326   & 8119  & 3327  & 0.388      & 0.514  \\ \hline \hline

\Padding       & (4, 4096, 16)  & 312   & 17639 & 1531  & 0.410      & 0.508  \\
\GroupByLength & (4, 4096, 16)  & 312   & 8621  & 3133  & 0.535      & 0.506  \\
\PaddingFree   & (4, 4096, 16)  & 312   & 8380  & 3223  & 0.410      & 0.508  \\ \hline
\Packing       & (4, 4096, 16)  & 103   & 7984  & 3383  & 0.441      & 0.582  \\
\Packing       & (1, 16384, 16) & 103   & 9919  & 2722  & 0.408      & 0.594  \\
\PackingPosId  & (4, 4096, 16)  & 103   & 7647  & 3532  & 0.435      & 0.542  \\
\PackingPosId  & (1, 16384, 16) & 103   & 7814  & 3455  & 0.403      & 0.541  \\
\Multipack     & (1, 16384, 16) & 112   & 7909  & 3415  & 0.392      & 0.539  \\
\Sortedpack    & (1, 16384, 16) & 107   & 7850  & 3441  & 0.394      & 0.540  \\
\Randompack    & (1, 16384, 16) & 112   & 7954  & 3396  & 0.393      & 0.539  \\ \hline
\Packing       & (4, 4096, 6)   & 274   & 8028  & 3365  & 0.435      & 0.574  \\
\Packing       & (1, 16384, 6)  & 274   & 9958  & 2711  & 0.402      & 0.593  \\
\PackingPosId  & (4, 4096, 6)   & 274   & 7647  & 3532  & 0.432      & 0.516  \\
\PackingPosId  & (1, 16384, 6)  & 274   & 8036  & 3360  & 0.400      & 0.515  \\
\Multipack     & (1, 16384, 6)  & 299   & 7849  & 3442  & 0.389      & 0.514  \\
\Sortedpack    & (1, 16384, 6)  & 286   & 7892  & 3422  & 0.391      & 0.518  \\
\Randompack    & (1, 16384, 6)  & 299   & 7986  & 3382  & 0.389      & 0.515  \\ \hline
\end{tabular}
\vspace{3mm}
\caption{Finetuning CodeLlama-Python-7B for 1 epoch on the Stack\_20k dataset: performance comparison of different sample packing approaches. The first group in each section uses minibatches and the second and third groups use offline packing.  Throughput   (train tokens/second) is improved substantially using  Packing  with Position IDs. Loss performance is the same for padding and  minibatch packing.}
\label{tab:stack_20k}
\end{table}

%%%====================
\section{ Conclusions } 
%%%====================

We provide a lightweight solution for packing with PositionIDs that has been integrated into Hugging Face Transformers library version 4.44 and we demonstrate its  benefits.
%%%=========== References ==========%%%
\clearpage
\bibliographystyle{plainnat}
\bibliography{packing}

\end{document}